\documentclass[10pt,twocolumn,letterpaper]{article}

\usepackage{3dv}
\usepackage{times}
\usepackage{epsfig}
\usepackage{graphicx}
\usepackage{amsmath}
\usepackage{amssymb}
\usepackage{xcolor}
\usepackage{booktabs}

\usepackage[pagebackref=true,breaklinks=true,letterpaper=true,colorlinks,bookmarks=false]{hyperref}
\newcommand{\tablestyle}[2]{\setlength{\tabcolsep}{#1}\renewcommand{\arraystretch}{#2}\centering\footnotesize}
\usepackage{cuted}
\usepackage{capt-of}

\threedvfinalcopy 

\def\threedvPaperID{8} 

\ifthreedvfinal\fi
\begin{document}

\title{4D Human Body Capture from Egocentric Video via 3D Scene Grounding}

\author{
  Miao Liu$^{1}$\thanks{This work was done when M. Liu was at ETH Z\"{u}rich.} ,\; Dexin Yang$^{3}$,\;  Yan Zhang$^{3}$,\; Zhaopeng Cui$^{2}$,\; James M. Rehg$^{1}$,\; Siyu Tang$^{3}$ \\
  $^1$ Georgia Institute of Technology, Atlanta, United States\\
  $^2$ Zhejiang University, Hangzhou, China \\
  $^3$ ETH Z\"{u}rich, Switzerland 
}

\maketitle
\thispagestyle{empty}

\begin{strip}\centering
\includegraphics[width=1.0\linewidth]{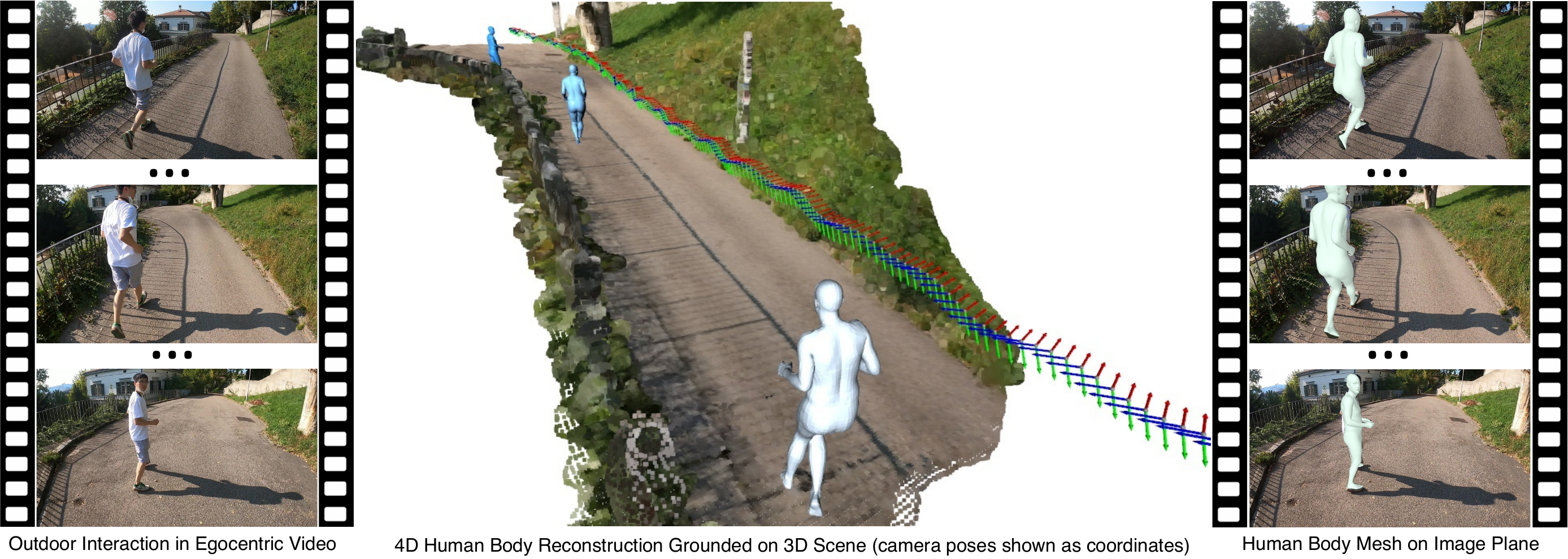}

\captionof{figure}{As shown in the middle figure, we seek to reconstruct 4D second-person human body meshes that are grounded on the 3D scene captured in an egocentric view. Our method exploits 2D observations from the entire video sequence and the 3D scene context to optimize human body models over time, and thereby leads to more accurate human motion capture and more realistic human-scene interaction.
\label{fig:teaser}}
\end{strip}
\begin{abstract}
We introduce a novel task of reconstructing a time series of second-person\footnote{We denote the social partner of the camera-wearer as ``second-person" throughout the paper. The same notation was also adopted in~\cite{ng2020you2me,yonetani2016recognizing}} 3D human body meshes from monocular egocentric videos. The unique viewpoint and rapid embodied camera motion of egocentric videos raise additional technical barriers for human body capture. To address those challenges, we propose a simple yet effective optimization-based approach that leverages 2D observations of the entire video sequence and human-scene interaction constraint to estimate second-person human poses, shapes, and global motion that are grounded on the 3D environment captured from the egocentric view. We conduct detailed ablation studies to validate our design choice. Moreover, we compare our method with the previous state-of-the-art method on human motion capture from monocular video, and show that our method estimates more accurate human-body poses and shapes under the challenging egocentric setting. In addition, we demonstrate that our approach produces more realistic human-scene interaction. 
\end{abstract}

\section{Introduction}

Continuous advancements in the capabilities of Augmented Reality (AR) headsets promise new trends of entertainment, communication, healthcare, and worker productivity, and point towards a revolution in how we interact with the world and communicate with each other. Egocentric vision is a key building block for these emerging capabilities, as AR experiences can benefit from an accurate understanding of the user's perception, attention, and actions. Substantial progress has been made in understanding human-object interaction~\cite{poleg2016compact,fathi2011understanding,Damen2018EPICKITCHENS,li2015delving,Li_2018_ECCV,furnari2019rulstm,li2020eye,liu2019forecasting,ego-topo} and social interaction~\cite{ye2015detecting,fathi2012social,soo2015social,yonetani2016recognizing,yagi2018future} from egocentric videos. However, future intelligent AR headsets should have the ability to capture the subtle nuances of second-person body pose and render an interactive 3D avatar that is grounded in the 3D scene as it is captured from an egocentric point of view. To this end, we introduce a novel task of 4D second-person full body capture from monocular egocentric videos. As shown in Fig.~\ref{fig:teaser}, we seek to reconstruct a time series of 3D second-person body meshes that are temporally-consistent and grounded on the reconstructed 3D scene.

3D human body capture from videos is a key challenge in computer vision, which has received substantial attention over the years~\cite{Kanazawa_2019_CVPR,VIBE:CVPR:2020,MuVS:3DV:2017,von2018recovering}. However, none of the previous works considered the challenging setting of reconstructing 3D second-person human body from an egocentric video\footnote{We note that another branch of prior work addresses the related but quite different task of predicting the 3D body pose of the \emph{camera-wearer} from egocentric video~\cite{ng2020you2me,jiang2017seeing,yuan20183d,tome2019xr,HPS}.}. The unique viewpoints and embodied camera motions of egocentric video create formidable technical obstacles to 3D body estimation, causing previous SOTA methods for video-based motion capture to fail. For example, the close interpersonal distances that characterize social interactions result in partial observation of the second-person as body parts move in and out of the frame. The egocentric camera motion creates an additional set of challenges, as the second-person motion is entangled with the embodied movement of the camera wearer. 

To address these challenges, we propose a simple yet effective optimization-based method that jointly considers a time series of 2D observations and 3D scene information. Our key insight is that the 3D scene provides additional evidence for estimating partially observable human body models. Previous work~\cite{PROX:2019} used a 3D scanner to obtain high quality 3D scene reconstructions, however this approach is not scalable, and is infeasible for outdoor egocentric capture settings. In contrast, we are the first to show that Structure-from-Motion (SfM) can provide a valuable 3D scene context for partially observable body estimation. This is particularly challenging because SfM estimates are up to an unknown scale, and directly placing the 3D body meshes into the reconstructed 3D scene and enforcing human-scene contact will result in unrealistic human-scene interaction. To overcome this challenge, we carefully design the optimization method so that it not only encourages human-scene contact, but also estimates the relative scale between 3D human body and scene reconstruction.\footnote{The prior work~\cite{PROX:2019} did not face the challenge of scale ambiguity because their 3D scene models came from a 3D scanner.} We further unite the time series of body models with a temporal prior to recover more plausible \emph{global human motion} even when the second-person body captured by the egocentric view is only partially observable.

Because existing egocentric datasets were not collected to address the problem of reconstructing the second-person body pose and shape in 4D, we have collected a new egocentric video dataset -- \emph{EgoMoCap}. Interactions between the first- and second-person in EgoMoCap were structured to yield a variety of interactions over a variety of interpersonal distances, which efficiently cover the variability in real-world social interactions in a compact number of clips. In contrast, previous social interaction datasets such as~\cite{fathi2012social} are naturalistic, but do not systematically cover the range of interpersonal distances needed for research in 4D capture. EgoMoCap is annotated with 2D human keypoints at the frame level. Using EgoMoCap, we compare our body capture approach with the previous state-of-the-art methods for human motion capture from monocular videos, and demonstrate that our method can address the challenging cases where the second-person human body is partially observable. Moreover, we demonstrate that our method can solve the relative scale between 3D scene reconstruction and 3D human body reconstruction from monocular videos, and thereby produce more realistic human-scene interactions. Detailed ablation studies highlight the benefits of our method. In summary, our work makes the following contributions:

\noindent \textbullet\ We introduce a novel problem of reconstructing time series of second-person poses and shapes from egocentric videos. To the best of our knowledge, we are the first to capture global human motion grounded in the 3D scene.

\noindent \textbullet\  We propose a simple yet effective optimization-based approach that jointly considers a time series of 2D observations and 3D scene context for accurate 4D human body capture. In addition, our approach addresses the scale ambiguity of 3D reconstruction from monocular videos.

\noindent \textbullet\  We conduct detailed experiments on our novel EgoMoCap dataset and show that our approach can more accurately reconstruct second-person human body, and encourage more realistic human-scene interaction.

\section{Related Work}
The most relevant works to ours are prior investigations of 4D human body reconstruction and human-scene interaction modeling. Our work is also related to recent efforts on reasoning about social interaction from egocentric videos. Specifically, we compare our EgoMoCap dataset with other egocentric human interaction datasets. 

\noindent  \textbf{4D Human Body Reconstruction}.\ 
A rich literature has addressed the topic of human body reconstruction. Previous approaches~\cite{Bogo:ECCV:2016, SMPL-X:2019, kolotouros2019learning,kanazawa2018end,anguelov2005scape,martinez2017simple,sun2017compositional,romero2017embodied} have demonstrated great success in inferring 3D human pose and shape from a single image. More closely-related to this work are prior efforts that leverage video of a moving peprson to infer a time series of 3D human body poses and shapes. Alldieck et al.\ \cite{alldieck2017optical} used optical flow to estimate temporally-coherent human bodies from monocular videos. Tung et al.\ \cite{tung2017self} introduced a self-supervised learning method that uses optical flow, silhouettes, and keypoints to estimate SMPL human body parameters from two consecutive video frames. Dabral et al.\ \cite{dabral2018learning} presented a weakly supervised learning framework for learning 3D human body pose, and adopted a temporal network to harmonize sequences of 3D pose estimations.  ~\cite{kanazawa2019learning,pavllo20193d} used a fully convolutional network to predict 3D human pose from 2D images sequences. Kocabas et al.~\cite{VIBE:CVPR:2020} proposed an adversarial learning framework to produce realistic and accurate human pose and motion from video sequences. Tripathi et al.\ \cite{tripathi2020posenet3d} also explored the knowledge distillation method for 3D human pose prediction. Shimada et al.\ \cite{PhysCapTOG2020} used a physics engine to capture physically plausible and temporally stable 3D human motion. Those previous works \emph{assumed a fixed camera view and fully observable human body.} Two works \emph{separately} considered either the partial observation setting or moving camera setting for human body reconstruction. Rockwell et al.\ \cite{Rockwell2020} proposed a self-training pipeline for reconstructing 3D human body poses from truncated single image frames. Huang et al.~\cite{MuVS:3DV:2017} enforced temporal coherence to reconstruct body pose from monocular videos with a moving camera. In contrast, we are the first to tackle the challenging task of reconstructing time-varying body models from egocentric video characterized by both partial observability and significant camera motion.

Other prior works addressed the question of body reconstruction from moving cameras by either combining video with IMU sensor data or leveraging multiple cameras~\cite{von2018recovering,MuVS:3DV:2017,wang2017outdoor}. \cite{von2018recovering} jointly optimize the camera pose and human body model by leveraging IMU sensory data.\  Wang et al.\ \cite{wang2017outdoor} proposed to utilize multiple cameras for outdoor human motion capture. More recently, Guzov et al.~\cite{HPS} proposed to estimate the full 3D human pose and location of the camera-wearer within large 3D scenes, by means of wearable sensors. In contrast to these works, we seek to estimate the second-person human motion grounded in the 3D scene using \emph{only monocular egocentric videos}.

\noindent \textbf{Human-Scene Interaction}.\ Several prior works on human-scene interaction seek to reason about environment affordance\ \cite{ego-topo,gupta20113d,grabner2011makes,delaitre2012scene,wang2017binge,koppula2015anticipating,chen2018subjects,nagarajan2018grounded,liu2021egocentric}. Our work is more relevant to previous efforts on using the environmental cues to estimate 3D human body models. Savva et al.\ \cite{savva2016pigraphs} proposed to learn a probabilistic model that captures how humans interact with the indoor scene from RGB-D sensors. Li et al.\ \cite{li2019estimating} factorized estimating 3D person-object interactions into an optimal control problem, and used contact constraints to recover human motion and contact forces from monocular videos. Zhang et al.~\cite{zhang2020phosa} proposed an optimization-based framework that incorporates the scale loss to jointly reconstruct the 3D spatial arrangement and shape of humans and objects in the scene from a single image. Zhang et al.\ \cite{zhang2020generatingnew,zhang2020generating} studied the problem of generating plausible human bodies grounded in 3D scenes. The most relevant work is PROX, from Hassan et al.~\cite{PROX:2019}, which introduced an optimization based method to use the 3D scene context for estimating more accurate human pose and shape from a single image. However, our work differs from PROX on several important aspects: First, PROX uses the Matterport 3D scanner to pre-scan the 3D scenes, whereas we use the Structure from Motion (SfM) to reconstruct the 3D scene geometry, resulting in a more general and scalable solution for 4D human pose and shape capture from a single egocentric camera. Second, we show that our contact term can address the challenging problem of the relative scale ambiguity between the estimated 3D human body and the reconstructed 3D scene from monocular videos, whereas the contact term used in PROX is only used to improve the physical plausibility. Third, the problem setting in PROX assumes a static camera with known 3D scene geometry, whereas we capture human motion in unconstrained environments with a moving egocentric camera, resulting in truncation and severe ego-motion.

\noindent \textbf{Egocentric Social Interaction}.\ 
Understanding human social interaction has been the subject of many recent efforts in egocentric vision~\cite{ye2015detecting,soo2015social,fathi2012social,yagi2018future,yonetani2016recognizing,yonetani2016recognizing}. Several egocentric datasets have been proposed for the analysis of human social behavior. The NUS Dataset\ \cite{narayan2014action} and JPL Dataset\ \cite{ryoo2013first} support more general human interaction classification tasks. Yonetani et al.\ \cite{yonetani2016recognizing} collected a paired egocentric human interaction dataset to study human action and reaction. Park et al.\ \cite{soo2016egocentric} introduced an RGB-D egocentric dataset -- EgoMotion, for forecasting first-person walking trajectory. Ng et al. intorduced You2Me dataset\ \cite{ng2020you2me} to study the problem of egocentric body pose prediction. Fathi et al.~\cite{fathi2012social} presented an egocentric video dataset that captures conversational interactions within a social group, and therefore limited second-person motion is captured in this dataset. Recently, the Ego4D social dataset from~\cite{GraumanEgo4D} was collected for understanding social communication behavior. In fact, none of those datasets were designed to study the \emph{second-person body pose} and \emph{human-scene interaction}. In prior datasets, the majority of captured second-person bodies are either largely occluded by objects or frequently truncated by the frustum, which makes their utilization for full body capture infeasible. In contrast, our EgoMoCap dataset focuses on outdoor social interaction scenarios where the second-person body has less occlusion and the Structure-from-Motion is more robust.

\begin{figure*}[t]
\centering
\includegraphics[width=0.95\linewidth]{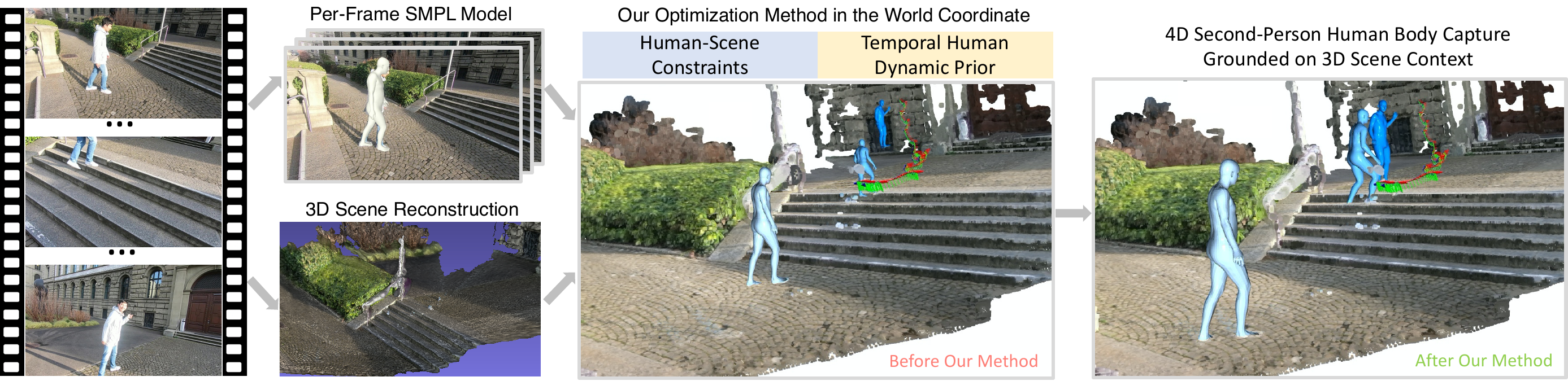}
\caption{Overview of our method. We first reconstruct the 3D scene $M_s$ using Structure-from-Motion, and estimate the SMPL-X body model $M_b$ from each video frame based on 2D observation. We then transform the human bodies from local image coordinates into 3D world coordinates. Furthermore, we use the human-scene contact term $E_C$ and temporal human dynamic prior $E_T$ to reconstruct the time series of 4D human body poses and shapes that are grounded on the 3D environment. Our proposed method not only encourages more realistic human-scene interaction by resolving the relative scale between 3D scene reconstruction and 3D human body reconstruction from monocular videos, but also addresses challenging cases where the second-person human body is partially observable.} \vspace{-1.5em}
\label{fig:overview}
\end{figure*}

\section{Method}
We denote an input monocular egocentric video as $x=(x^1, ..., x^t)$ with its frame $x^t$ indexed by time $t$. We estimate the human body pose and shape at each time step from input $x$. Due to the unique viewpoint of egocentric video, the captured second-person body is partially observable within a time window. In addition, the second-person body motion is entangled with the camera motion, creating additional barriers for the enforcement of temporal coherency. To address these challenges, we propose an optimization method that jointly considers the 2D observations of the entire video sequence and the 3D scene context in order to more accurately reconstruct the 4D human body in the presence of partial observability. We illustrate our method in Fig.~\ref{fig:overview}. Specifically, we first recover the 3D human body at each time instant from the 2D observation of $x^t$. We then estimate Structure from Motion~\cite{schoenberger2016sfm} (SfM) to project a sequence of 3D body meshes into 3D world coordinates based on the recovered global camera motion, and further adopt a contact term to enforce appropriate human-scene interaction. In addition, we combine the 2D cues from entire video sequences for reconstructing temporally-coherent time series of body poses using a human dynamics prior. In the following sections, we introduce each component of our method.

\subsection{Human Body Model Estimation}
We use the differentiable human body model SMPL-X~\cite{SMPL-X:2019} to represent the body, hands, and facial expression. SMPL-X produces a body mesh of a fixed topology with $N_b=10475$ vertices, using a compact set of body configuration parameters. Specifically, the shape parameter $\beta$ encodes variations in height, volume, and body proportions; $\theta$ encodes the 3D body pose, hand pose, and facial expression information; and $\gamma$ denotes the body translation. Formally, the SMPL-X function is defined as $M_b(\beta,\theta,\gamma)$. It outputs a 3D body mesh $M_b = (V_b,F_b)$, where $V_b \in R^{N_b\times3}$ and $F_b$ denote the body vertices and triangular faces. Similar to~\cite{SMPL-X:2019,Bogo:ECCV:2016}, we factorize fitting the SMPL-X model to each video frame as an optimization problem. Formally, we optimize $(\beta,\theta,\gamma)$ by minimizing:
\begin{align}
E_M(\beta,\theta,\gamma,K,J_{est}) & = \nonumber E_J(\beta,\theta,\gamma,K,J_{est}) 
\\ &+ \lambda_{\beta}E_{\beta}(\beta) + \lambda_{\theta}E_{\theta}(\theta),
\label{eq:}
\end{align}
\noindent where K is the intrinsic camera parameters; the shape prior term $E_{\beta}(\beta)$ is learned from SMPL-X model body shape training data and the pose prior term $E_{\theta}(\theta)$ is learned from CMU MoCap dataset~\cite{cmumocap}; $ \lambda_{\beta}$ and $\lambda_{\theta}$ denote the weights of $E_{\beta}(\beta)$ and $E_{\theta}(\theta)$; $E_J$ refers to the energy function that minimizes the weighted robust distance between the 2D projection of the body joints, hand joints and face landmarks, and the corresponding 2D joints estimation from OpenPose~\cite{cao2017realtime,wei2016cpm}. $E_J$ is given by:
{\small
\begin{align}
& E_J(\beta,\theta,\gamma,L,J_{est}) \nonumber \\ &= 
\sum_{joint\:i} k_iw_i\rho_J(\Pi_K(R_{\theta\gamma}(J^i(\beta))-J_{est}^i),
\end{align}}%
where $J(.)$ returns 3D joints location based on embedded shape parameters $\beta$, and $R_{\theta\gamma}(.)$ transforms the joints along the kinematic tree according to the pose $\theta$ and body translation $\gamma$; $\Pi_K$ is the 3D to 2D projection function based on intrinsic parameters $K$; $J_{est}$ refers to the 2D joints estimated from OpenPose; $w_i$ is the 2D joints detection confident score which accounts for the noise in 2D joint estimation; $k_i$ is the per-joint weights for annealed optimization as in~\cite{SMPL-X:2019}; $\rho_J$ denotes a robust Geman-McClure error function~\cite{geman1987statistical} that downweights outliers, which is given by: $\rho_J(e)= e^2/(\sigma_J^2 + e^2)$, where $e$ is the residual error, and $\sigma_j$ is the robustness constant, which is chosen empirically.

\subsection{Egocentric Camera Representation}
To capture 4D second-person bodies that are grounded on the 3D scene from egocentric videos, we need to take the embodied camera motion into consideration. Here we elaborate the egocentric camera representation adopted in our method. Formally, we denote $T_{cb} \in R^{4\times4}$ as the transformation from the human body coordinate to the egocentric camera coordinate, and $T_{wc}$ as the transformation from the egocentric camera coordinate to the world coordinate. Note that $T_{cb} \in R^{4\times4}$ is derived from the translation parameter $\gamma$ of SMPL-X model fitting introduced in Sec. 3.1, while $T_{wc}$ is returned from COLMAP Structure from Motion (SfM)~\cite{schoenberger2016sfm}. In order to utilize the 3D scene context and enforce the temporal coherency on reconstructed human body meshes, we project the 3D second-person body vertices $V_b$ into world coordinate using human body to world transformation $T_{wb}$, which is given by:
{\small
\begin{align}
\label{eq:transformation}
\small
\hat{V}_{wb}^t &= T_{wb}^t \hat{V}_b^t=T_{wc}^t T_{cb}^t \hat{V}_b^t,
\end{align}}%
\noindent where $\hat{V}_b^t$ refers to the body vertices at time step $t$, represented in homogeneous coordinate.

\subsection{Optimization with 3D Scene}
\noindent\textbf{3D Scene Representation}.\ The structure of the 3D scene constrains and informs human behavior, and therefore 3D scene context can play an important role in 3D human body recovery. Human-scene interaction can be described in relation to 3D surfaces, and therefore we adopt a mesh representation for the 3D scene. Formally, we denote the 3D scene mesh as $M_s=(V_s,F_s)$, where $V_s\in R^{N_s\times3}$ denotes the vertices of the scene representation, and $F_s$ denotes the corresponding triangular faces. We use the dense environment reconstruction from COLMAP to obtain $M_s$. Specifically, COLMAP first reconstructs a sparse representation of the scene and the camera poses of the input images using SfM. It then calculates the depth and normal maps for all registered images, and fuses the depth and normal maps into a dense point cloud with normal information. Finally, Poisson Reconstruction is used to generate the 3D scene mesh representation $M_s$.

\noindent\textbf{Human-Scene Contact}.\ Note that the reconstructed 3D scene from the monocular video is up to a scale. To address this scale ambiguity, we design a novel energy function that not only encourages contact between the human body and 3D scene, but also estimates the relative scale between 3D scene mesh $M_s$ and 3D body mesh $M_b$. Specifically, we make use of the annotation from ~\cite{PROX:2019}, where a candidate set of SMPL-X mesh vertices $V_c \in V_b$  to contact with the world were provided. We then multiply an optimizable scale parameter $S \in R$ to human body vertices $V_c$ during optimization. Therefore, the energy function for enforcing human-scene contact is given by:
{\small
\begin{align}
&E_C(\beta,\theta,\gamma,V_s,S) = \\ \nonumber
&\sum_{i=1}^{t}\sum_{v_c\in V_c^t} \rho_c(\min_{v_s\in V_s}||T_{wb}^t (Sv_c)-v_s||),
\end{align}}
\noindent where $\rho_c$ is the robust Geman-McClure error function introduced before, and $T_{wb}$ is human body to world transformation introduced in Eq.~\ref{eq:transformation}. Note that the scale factor $S$ is shared across the video sequence. This is because we estimate a consistent 3D shape parameter $\theta$ from the entire sequence by taking the median of all the shape parameters obtained from the per-frame SMPL-X model fitting. 

\subsection{Human Dynamics Prior}
Fitting SMPL-X human body model to each video frame will incur notable temporal inconsistency. Due to drastic camera motion, this problem is further amplified under egocentric scenarios. Here, we propose to use the empirical human dynamics priors to enforce temporal coherency on human body models in the world coordinates. Formally, we have the following energy function:
{\small
\begin{align}
\label{eq:motion}
&E_T(\beta,\theta,\gamma) = \\ \nonumber
&\sum_{i=2}^{t}\sum_{V}(1-w_v)\rho_T((V_{wb}^{i+1}-V_{wb}^i)-(V_{wb}^i-V_{wb}^{i-1})),
\end{align}}%
where $V_{wb}^i$ is the 3D human body vertices at time step $i$, transformed in world coordinate as in Eq.\ \ref{eq:transformation}; $\rho_T$ is another robust Geman-McClure error function that accounts for possible outliers; and $w_v^i$ is the confidence score of 2D human keypoints estimation. As shown in Eq.~\ref{eq:motion}, we design this energy function to focus on body parts that do not have reliable 2D observations (caused by the unique egocentric viewpoint). Notably, we assume a zero acceleration motion prior. This naive prior was proven to be effective in capturing human motion in the outdoor environment~\cite{Arnab_2019_CVPR}.

\subsection{Optimization}
Putting everything together, we have the following energy function for our optimization method:
\begin{equation}
\small
E_{total} = \sum_{i=1}^{t}E_M^i+\lambda_C E_C+\lambda_T E_T,
\label{eq:full}
\end{equation}
where $E_M^i$ denotes the SMPL-X model fitting energy function for video frame $x^i$; $\lambda_C$ and $\lambda_T$ represent the weights for human-scene contact term and human dynamic prior term, respectively. We optimize Eq.\ \ref{eq:full} using a gradient-based optimizer Adam~\cite{kingma2014adam} w.r.t. SMPL-X body parameters $\beta,\theta,\gamma$, scale parameter $S$, and camera to world transformation $T_{wc}$. Note that the SfM already provides an initialization of $T_{wc}$, and incorporating $T_{wc}$ into the optimization can further smooth the global second-person human motion.

Note that $E_M$ performs model fitting at each time step, while $E_C$ and $E_T$ optimize a time series of body models. In addition, $E_C$ and $E_T$ seek to optimize human body parameters in world coordinates, and the scale ambiguity can  cause the gradients of the contact term to shift the body global position in the wrong direction. Therefore, we propose a multi-stage optimization strategy. Specifically, we set $\lambda_C$ and $\lambda_T$ to be zero, so that the optimizer will only look at the 2D observations in the first stage.  We then set $\lambda_C$ to be 0.1, keeping $\lambda_T$ as zero, and freezing $T_{wc}$, so that the optimizer will focus on recovering the scale parameter $S$. At the final stage, we set $\lambda_T$ to 0.1 and enable the gradients of $T_{wc}$ to enforce temporal coherency.

\section{Experiments}
\subsection{Dataset and Metrics}
\noindent \textbf{Datasets}.\ To study the problem of second-person human body reconstruction, we present a new egocentric social interaction dataset -- EgoMoCap. This dataset consists
of 36 videos sequences from one-on-one interactions between 4 individuals. The camera wearer is equipped with a head-mounted camera, and the other participant is asked to interact with the camera wearer in a natural manner. The video sequences were recorded in 1920×1080 resolution at 60 fps using the GoPro Hero8 camera. This dataset captures 4 types of outdoor human social interactions: \emph{Greeting}, \emph{Touring}, \emph{Jogging Together}, and \emph{Throw and Catch}. We further annotate the captured second-person human bodies with 2D keypoints via Amazon Mechanical Turk (AMT).

\noindent \textbf{Evaluation Metrics}.\ For our experiments, we evaluate the human body reconstruction accuracy, motion smoothness, and the plausibility of the human-scene interaction.

\noindent \textbullet\ \textbf{Human Body Reconstruction Accuracy}:\ We acknowledge that the 3D ground truth of human bodies can be obtained from RGB-D data~\cite{PROX:2019}, or Motion Capture Systems~\cite{SIP,mahmood2019amass}. However, capturing the 3D human body ground truth in naturalistic outdoor social interaction setting remains a challenge. Therefore, we follow~\cite{yuan20183d,Rockwell2020} to evaluate the reconstruction quality using per-joint 2D projection error (PJE) on the image plane. Specifically, we report PJE-P on frames with partially observable second-person body, and PJE-U on frames with mostly untruncated second-person body (uniform sampled frames). Here, we evaluate human body poses, even though our method has the capacity of reconstructing 3D hands and faces, as human-scene contact used in our method has minor influence on facial expression and hand pose during social interaction.

\noindent \textbullet\ \textbf{Motion Smoothness}:\ We follow~\cite{yuan20183d} to adopt a physics-based metric that uses average magnitude of joint accelerations to measure the smoothness of the estimated pose sequence.
Thus, a lower value indicates that the times series of body meshes have more consistent human motion. Note that the motion smoothness is evaluated on 3D human joints projected in world coordinate. For fair comparison, we normalize the scale factor when reporting the results.

\noindent \textbullet\ \textbf{Plausibility of Human-Scene Interaction}:\
To evaluate whether our method provides more realistic human-scene interaction, we transform the human body meshes into 3D world coordinates, render the results as video sequences, and further upload them to Amazon Mechanical Turk (AMT) for a user study. Specifically, we put the rendered results of all compared methods and our method side-by-side (sample videos can be found in supplementary materials), and ask the AMT worker to choose the instance that has the most realistic human-scene interaction.

\subsection{Quantitative Results}
In this section, we introduce our quantitative experimental results. We first present detailed ablation studies, and then compare our method with state-of-the-art for 3D human body reconstruction from monocular videos.

\begin{table}

\setlength{\tabcolsep}{2.7pt} 
\renewcommand{\arraystretch}{1.05} 
\tablestyle{2pt}{1.0}
\centering
\begin{tabular}{c|cccc}
\toprule
\textbf{}
Method      & PJE-P $\downarrow$  &PJE-U $\downarrow$  & Smoothness $\downarrow$   & User Study $\uparrow$      \\ \hline 
$E_M$  (SMPLify-X)                         &73.14 &\textbf{22.19}  &5.33 &7.4   \\ 
$E_M$+$E_C$                          &87.74 &30.09 &5.72 &23.2   \\ 
$E_M$+$E_T$                   & 75.14 &23.93 &2.23 &13.7  \\ 
$E_M$+$E_C$+$E_T$ (Ours) &\textbf{66.03} & 24.03   &\textbf{1.82} &\textbf{55.7}   \\ 
\toprule
\end{tabular}
\caption{Ablation study for our proposed method. We analyze the role of human dynamic prior $E_T$ and human-scene interaction term $E_C$. Note that PJE-P is the core metric to evaluate whether the method can recover more accurate human body poses and shapes of partially observable second-person human body. ($\uparrow$/$\downarrow$ indicates higher/lower is better)
}\vspace{-1.8em}
\label{table:ablationstudy}
\end{table}

\begin{figure*}[t]
\centering
\includegraphics[width=0.96\linewidth]{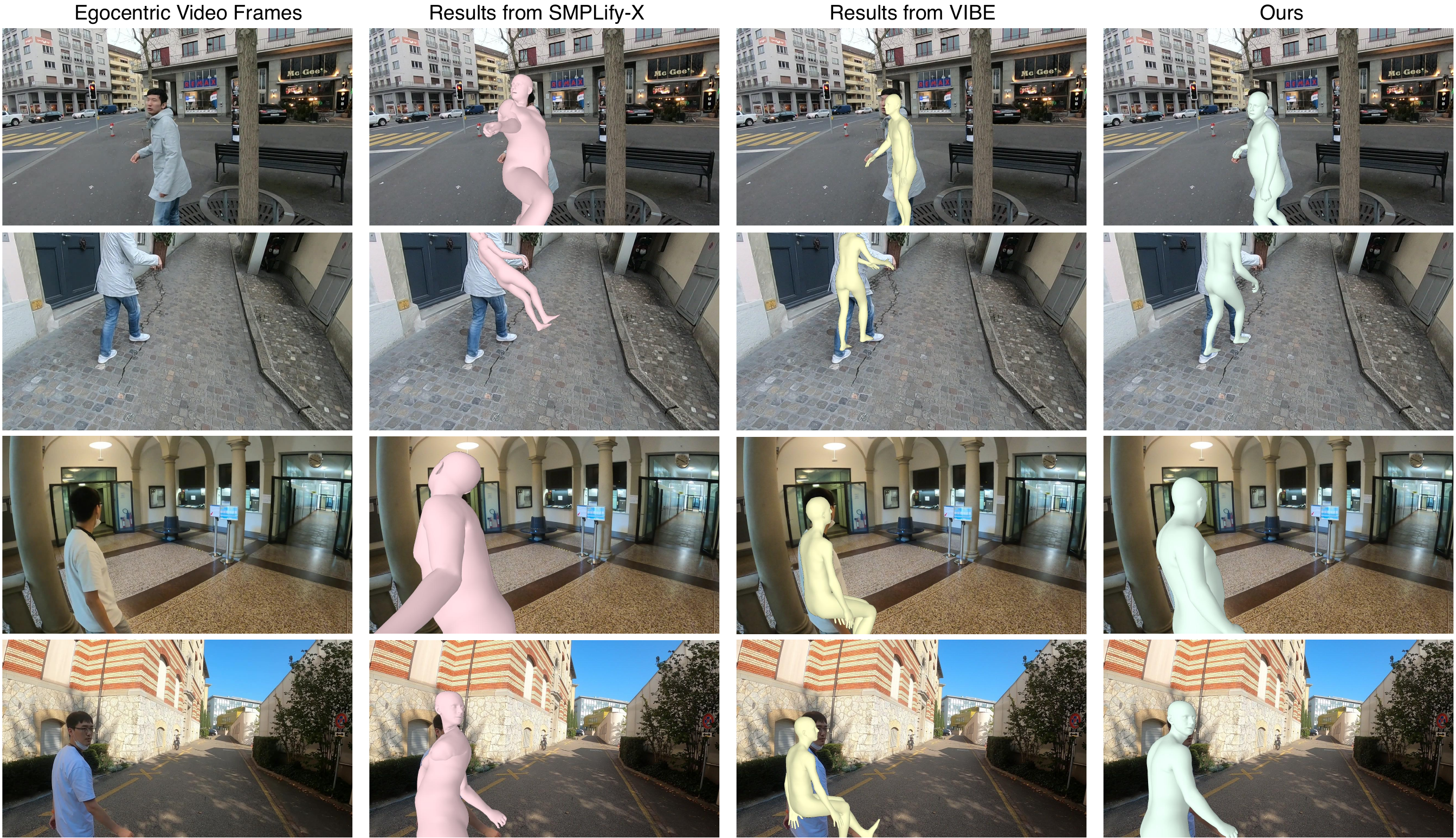}
\caption{Qualitative comparison between our method and other approaches. The first column is the original video frames;  the second column is the results from SMPlify-X, the third column is the results from VIBE, and the last shows our results. Our approach can address the challenging cases when the second-person body is partially observable. } \vspace{-1.2em}
\label{fig:vis}
\end{figure*}

\begin{figure*}[t]
\centering
\includegraphics[width=0.96\linewidth]{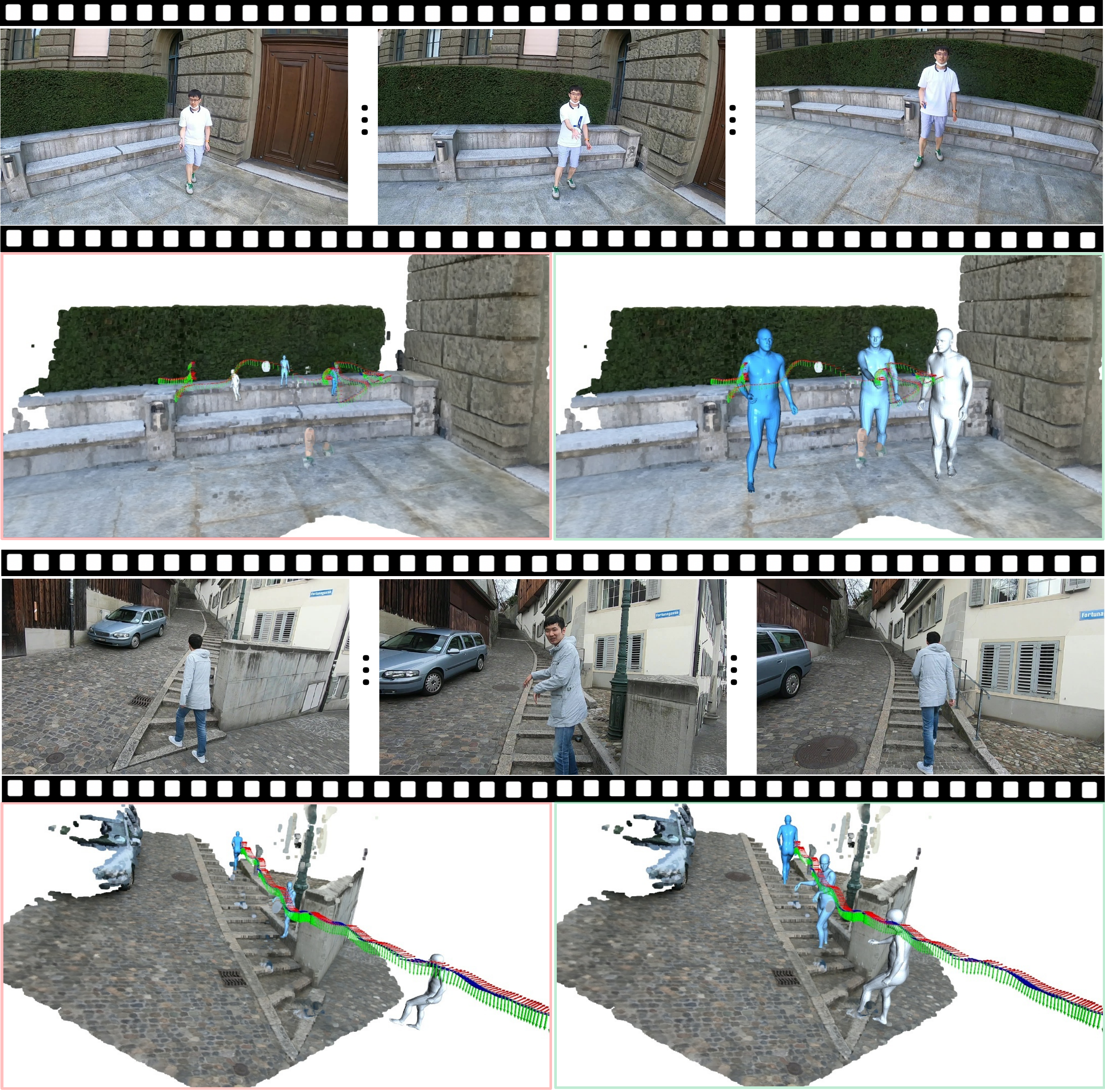}
\caption{Visualization of time series of human bodies in the world coordinate. We visualize both results of SMPLif-X baseline (Left) and our method (Right) projected into 3D scene reconstruction.  Our method recovers the scale ambiguity between 3D scene reconstruction and 3D body reconstruction from monocular video, and therefore leads to more plausible human-scene interaction.}
\label{fig:vis_scale}\vspace{-1.6em}
\end{figure*}

\noindent \textbf{Ablation Study}. We first analyze the functionality of the terms in Eq.~\ref{eq:full}. The results are summarized in Table~\ref{table:ablationstudy}. $E_M$ refers to the baseline method that performs per-frame fitting with 2D observations as in SMPLify-X~\cite{SMPL-X:2019}. $E_M$ has undesirable performance on PJE-P, motion smoothness and human-scene interaction user study. In the second row $E_M+E_C$, we report the method that makes use of both human scene contact term and 2D observations. Though adding the contact term alone leads to more realistic human-scene interaction, it compromises the performance on 2D projection error and motion smoothness by a notable margin. $E_M+E_T$ in the third row refers to the method that optimizes the 2D observations together with the human dynamic prior term $E_T$. Not surprisingly, $E_T$ can significantly improve the motion smoothness. It is also worthy noting that using temporal prior alone can not improve the reconstruction quality. This suggests that simply enforcing temporal coherency without 3D scene grounding may cause the optimization method converging to sub-optimal equilibrium. In the last row, we present the results of our full optimization approach. Our method achieves the best performance on motion smoothness and plausibility of human-scene interaction. An interesting observation is that ours outperforms $E_M+E_T$ by a notable margin on motion smoothness. We speculate that this is because the physical human scene constraints narrow down the solution space of model fitting, and thereby leads to more optimal performance on temporal coherency. Note that, in the simple case when the figure is untruncated (PJE-U), $E_M$ gives good performance and slightly exceeds the accuracy of our full method. This is because PJE is a 2D metric, and therefore favors the method that adopts only 2D projection error as objective function during optimization. However, when the 2D observation can not be robustly estimated due to partial observation, our method outperforms other baselines by a significant margin (66.03 vs.\ 73.14 in PJE-P). Those results demonstrate that our method can address the challenge of partially observable human bodies, and estimate plausible global human motion grounded on the 3D scene.

\noindent \textbf{Comparison to VIBE}. Though many methods have addressed human body capture from monocular video. In Table~\ref{table:results}, we compare our approach with a widely-used competitive method -- VIBE~\cite{VIBE:CVPR:2020}. Since VIBE does not model the human-scene constraints, it provides unrealistic human-scene interaction. Moreover, the egocentric camera motion causes VIBE failing to capture coherent human motion. In contrast, our method outperforms VIBE on motion smoothness and human-scene interaction plausibility by a large margin. Though VIBE performs slightly better on PJE-U (22.45 vs. 24.03), it lags far behind of our method on PJE-P (75.91 vs. 66.03). We have to re-emphasize that the 2D projection error cannot reflect the true performance improvement of our method. This is because the 2D keypoints annotation is only available for visible human body parts, and therefore 2D PJE does not penalize the method that fits a wrong 3D body model to partially 2D observation. Take the VIBE result shown in the third row of Fig.~\ref{fig:vis} for an instance, the 2D projection error may have decent performance, yet the reconstructed 3D human body is completely wrong.
\begin{table}
\centering
\footnotesize
\begin{tabular}{c|cccc}
\toprule
Method      & PJE-P $\downarrow$  &PJE-U $\downarrow$ & Smoothness $\downarrow$     & User Study $\uparrow$    \\ \hline 
VIBE~\cite{VIBE:CVPR:2020}     & 75.91             &\textbf{22.45} &4.79 &17.2  \\ 
Ours                   &\textbf{66.03}  &24.03  &\textbf{1.85} &\textbf{82.8}\\ 
\bottomrule
\end{tabular}
\caption{Experiment results comparison with competitive method VIBE. ($\uparrow$/$\downarrow$ indicates higher/lower is better)}\vspace{-1.6em}
\label{table:results}\vspace{-0.5em}
\end{table}


\subsection{Discussion}

\noindent\textbf{Qualitative Results}.\ As shown in Fig.~\ref{fig:vis}, we first visualize our results by rendering the estimated body models from the egocentric viewpoint, so it can be directly imposed on the input video frames. Notably, both SMPLify-X and VIBE fail substantially for challenging cases where human body is partially-observable. Our method, on the other hand, makes uses of 3D scene context and harmonizes the 2D cues from the entire video sequence, and therefore successfully reconstructs the human body with only partial observation. We provide an additional visualization of the results of both $E_M$ baseline and our method in the world coordinate system in Fig.~\ref{fig:vis_scale}. By examining the SMPLify-X baseline results, we can observe an obvious mismatched scale between the 3D reconstruction of the human body and the 3D environment. In contrast, our method produces more plausible human body motion grounded in the 3D scene by resolving the relative scale between 3D scene reconstruction and 3D human body reconstruction from monocular videos. In the supplementary materials, we provide additional videos that  demonstrate the benefits of our approach.

\noindent\textbf{Limitations}.\ A key issue of our method is the need to retrieve the camera trajectory and 3D scene only from monocular RGB videos via Structure from Motion (SfM). Therefore, our method has the same bottleneck as SfM: Challenging factors such as dynamic scenes, featureless surfaces, large illumination change, etc., may cause visual feature matching to fail. One promising direction for future work is to incorporate additional sensing modalities (IMU, depth estimation, and multiple cameras) to further stabilize the 3D scene reconstruction in challenging conditions. Another issue is that the naive human motion prior (zero acceleration) adopted in our method may result in unrealistic motions in some cases. More efforts in learning motion priors could potentially address this issue. In summary, we believe our efforts constitute an important step forward for a largely unexplored egocentric vision task, and we hope our work can motivate the community to make further investments. 

\section{Conclusion}
We introduce a novel task of estimating a time series of 3D human body models for the second-person in an egocentric video, which are temporally-coherent and grounded on the 3D scene. To address the challenges of egocentric video, we propose an effective optimization-based method that exploits the 2D observations of the entire video sequence and human-scene contact for human motion capture. We conduct detailed experiments on our EgoMoCap dataset to demonstrate the benefits of our approach. We believe our work points to exciting research directions in egocentric social interaction analysis and 4D human body reconstruction.

\noindent \textbf{Acknowledgments}. Portions of this research were supported in part by National Science Foundation Award 2033413 and a gift from Facebook. 


{\small
\bibliographystyle{ieee_fullname}
\bibliography{egbib}
}

\end{document}